\documentclass{article} 
\usepackage{iclr2026_conference,times}

\usepackage{amsmath,amsfonts,bm}

\def\eqref#1{equation~\ref{#1}}

\def\1{\bm{1}}

\DeclareMathAlphabet{\mathsfit}{\encodingdefault}{\sfdefault}{m}{sl}
\SetMathAlphabet{\mathsfit}{bold}{\encodingdefault}{\sfdefault}{bx}{n}

\usepackage{hyperref}
\usepackage{url}
\usepackage{graphicx}
\usepackage{algorithm}
\usepackage{algorithmic}
\usepackage{amssymb}
\usepackage{amsmath}
\usepackage{multirow}
\usepackage{booktabs}
\usepackage{colortbl}
\usepackage[table]{xcolor}
\usepackage{tabularx}
\usepackage{placeins}

\definecolor{oursbg}{HTML}{F2F5F9}

\definecolor{bestred}{HTML}{C00000}
\definecolor{secondblue}{HTML}{0055D4}

\newcommand{\best}[1]{\textcolor{bestred}{\textbf{#1}}}
\newcommand{\second}[1]{\textcolor{secondblue}{#1}}

\definecolor{promptink}{HTML}{4B6691}
\newcommand{\prompttag}[1]{\textbf{\textcolor{promptink}{#1}}}

\title{Towards Active Cross-View Object Geo-Localization}

\author{
\textbf{Shunyu Yao}$^{1,*}$
\quad
\textbf{Xiaohan Zhang}$^{1,*,\dagger}$
\quad
\textbf{Zhuoran Yang}$^{1}$
\quad
\textbf{Haoqi Lai}$^{1}$
\\
\textbf{Qi Ming}$^{2}$
\quad
\textbf{Xiaoxi Hu}$^{3}$
\quad
\textbf{Hui-Liang Shen}$^{1}$
\quad
\textbf{Si-Yuan Cao}$^{4,5,\ddagger}$
\\[0.8em]
{\normalfont\small
$^{1}$College of Information Science and Electronic Engineering,
Zhejiang University, Hangzhou 310027, China
}\\
{\normalfont\small
$^{2}$College of Computer Science,
Beijing University of Technology, Beijing 100124, China
}\\
{\normalfont\small
$^{3}$State Key Laboratory of Intelligent Green Vehicle and Mobility,
Tsinghua University, Beijing 100084, China
}\\
{\normalfont\small
$^{4}$Ningbo Global Innovation Center,
Zhejiang University, Ningbo 315100, China
}\\
{\normalfont\small
$^{5}$Jinhua Institute of Zhejiang University,
Jinhua 321000, China
}
\\[0.5em]
{\normalfont\small
$^{*}$Equal contribution.
\quad
$^{\dagger}$Project lead.
\quad
$^{\ddagger}$Corresponding author.
}
\\[0.3em]
{\normalfont\small
\texttt{yaoshunyu306@gmail.com}
\quad
\texttt{zhangxh2023@zju.edu.cn}
\quad
\texttt{cao\_siyuan@zju.edu.cn}
}
}

\iclrfinalcopy


\begin{document}

\maketitle
\fancyhead{}

\begin{abstract}
Cross-view object geo-localization (CVOGL) typically assumes a fixed query image, overlooking the ability of mobile agents to actively acquire more informative observations. To address this limitation, we introduce \textbf{Active Cross-View Object Geo-Localization (ActiveGeo)}, where an agent sequentially selects new viewpoints and determines when to stop, aiming to improve localization with minimal observations. We further propose \textbf{ActiveMoPT}, an ActiveGeo framework with three-stage training. First, \emph{Multi-View Prompt-Preserving Adaptation} enables the model to aggregate multiple query views while reusing the initial prompt. Second, \emph{Trajectory-Guided Policy Initialization} uses supervised agent trajectories to learn viewpoint selection and initial stopping behavior. Third, \emph{Cost-Aware Policy Refinement} employs GRPO with a gain-cost reward to jointly optimize localization accuracy and observation efficiency. We also construct \textbf{ActiveGeo-858}, a zero-shot test set containing 858 scenes and 1,716 target annotations. Experiments show that ActiveMoPT achieves state-of-the-art performance on MoP-UAV using only 1.45 query views on average, and substantially outperforms previous CVOGL approaches under zero-shot evaluation on ActiveGeo-858. 
\end{abstract}

\section{Introduction}
\label{sec: Introduction}
Cross-view object geo-localization (CVOGL) aims to determine the geographic location of a specific object indicated by point prompts in a query image on the reference image \citep{DetGeo}. The query images can be captured from devices like phones, autonomous vehicles, and drones, while the reference images are typically satellite images. CVOGL is widely used in various applications, such as smart city management \citep{smart_city_1}, disaster monitoring \citep{disater_1, disater_2}, and robot navigation \citep{robot, robot2}. 

Despite the cross-view discrepancies in appearance, scale, and viewpoint \citep{DetGeo}, recent CVOGL approaches \citep{GeoFormer, TROGeo, HALO, ReCOT} have gradually matured and achieved promising localization performance. Meanwhile, the recent study \citep{mopt} has further extended CVOGL prompts from a single modality to flexible multimodal cues, substantially broadening the applicability of CVOGL in diverse interaction scenarios.

\begin{figure}[!htb]
  \centering
  \includegraphics[width=1.0\linewidth]{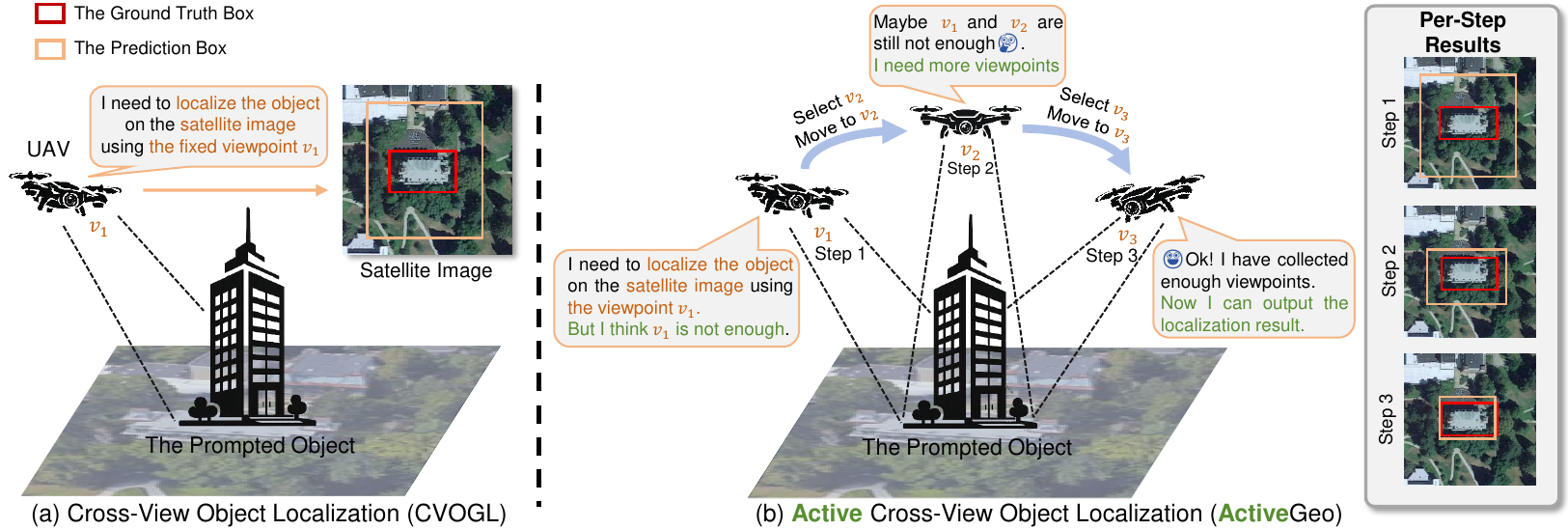}
  \vspace{-0.5cm}
   \caption{Comparison between previous passive cross-view object geo-localization (CVOGL) and our active cross-view object localization (ActiveGeo). (a) Conventional CVOGL localizes the target from a fixed query view. (b) ActiveGeo enables a UAV agent to sequentially acquire informative viewpoints and adaptively stop when sufficient observations have been collected. Please refer to zoomed-in view for better visualization.}
   \label{fig: head}
   \vspace{-0.2cm}
\end{figure}

However, existing CVOGL approaches \citep{OCGNet, GeoFormer, HALO, ReCOT} still follow a passive localization paradigm \citep{DetGeo}, where the query image is fixed and localization is performed from the given observation, as shown in Fig.~\ref{fig: head}(a). This overlooks the mobility of embodied agents, which can actively acquire new views when the current observation is occluded or otherwise unfavorable. Nevertheless, existing CVOGL works lack the ability to support such active view acquisition and decision-making.

Addressing this limitation is important for making CVOGL more consistent with real-world mobile-agent applications~\citep{robot, disater_1}. It also introduces agentic decision-making~\citep{AgentGYM-RL} into CVOGL, extending the task from passive prediction to active interaction and broadening its research scope. Therefore, we introduce \textbf{Active Cross-View Object Geo-Localization (ActiveGeo)}, where an agent sequentially selects new informative viewpoints and determines when to stop, with the goal of improving localization using minimal observations. UAVs are particularly suitable for ActiveGeo due to their high mobility and flexibility~\citep{university-1652}. We therefore focus on the UAV-based setting, as shown in Fig.~\ref{fig: head}(b).

Under this new formulation, a practical ActiveGeo should satisfy three essential requirements. (1) \emph{The model should support sequentially acquired query images while reusing the initial prompt}, since the target identity remains unchanged as the agent moves and repeated prompting is unrealistic. (2) \emph{The agent should select informative viewpoints based on its current localization state}, since unobserved views are unavailable before movement and additional observations may be redundant or noisy. (3) \emph{The agent should decide when to stop acquiring new views}, balancing potential localization gains against sensing, movement, and computation costs.

To meet these requirements, we build the \textbf{ActiveMoPT} upon MoPT \citep{mopt} through a three-stage training framework. (1) In \emph{Multi-View Prompt-Preserving Adaptation}, we introduce a multi-view adapter and gain-aware loss that enable MoPT to progressively aggregate multiple query observations while reusing the initial prompt. (2) In \emph{Trajectory-Guided Policy Initialization}, we construct agent trajectory data and perform supervised fine-tuning (SFT), allowing ActiveMoPT to
learn viewpoint selection and acquire an initial stopping capability. (3) In \emph{Cost-Aware Policy Refinement}, we optimize the policy through group relative policy optimization (GRPO) with a gain-cost reward that jointly considers localization gain and observation cost, leading to a better accuracy-efficiency trade-off. Besides, we further construct a dedicated zero-shot test set \textbf{ActiveGeo-858} beyond the existing MoP-UAV benchmark \citep{mopt} for ActiveGeo. It is built upon GeoText-1652 \citep{university-1652} and contains 858 scenes and 1,716 target annotations, with strict separation from MoP-UAV benchmark.  Extensive experimental results show that ActiveMoPT achieves state-of-the-art localization performance on the MoP-UAV Benchmark \citep{mopt} while using only 1.45 query views on average. Moreover, ActiveMoPT also substantially outperforms under zero-shot evaluation on ActiveGeo-858, demonstrating strong generalization to unseen scenes. Our main contributions are summarized as follows:

\begin{itemize}
    \item We introduce \textbf{Active Cross-View Object Geo-Localization
    (ActiveGeo)}, extending conventional CVOGL from passive prediction
    to active interaction. The new task requires an agent to actively acquire and select informative viewpoints, and determine when to stop under a limited observation budget.

    \item We construct \textbf{ActiveMoPT} through a three-stage training framework, consisting of \emph{Multi-View Prompt-Preserving Adaptation},
    \emph{Trajectory-Guided Policy Initialization}, and
    \emph{Cost-Aware Reinforcement Refinement}. 

    \item We construct \textbf{ActiveGeo-858}, a dedicated zero-shot test set
    containing 858 scenes and 1,716 target annotations for ActiveGeo. 
    
    \item Extensive experiments show that ActiveMoPT achieves state-of-the-art performance on MoP-UAV benchmark \citep{mopt} using only 1.45 query views on average, while also substantially outperforming previous approaches under zero-shot evaluation on ActiveGeo-858.
\end{itemize}


\section{Related Work}
\label{sec: Related Work}

\textbf{Cross-View Object Geo-Localization (CVOGL).}
CVOGL aims to localize a prompted object in a reference image from a cross-view query image. DetGeo~\citep{DetGeo} first formalizes this task, while subsequent approaches~\citep{VaGeo, TROGeo, OCGNet, ReCOT} further improve prompt representation, cross-view interaction, and localization decoding.

More recently, MoPT~\citep{mopt} broadens the prompting interface of CVOGL by allowing the target object to be specified using point, box, language, or their combinations. It encodes different prompt modalities and concatenates them into prompt tokens, which sequentially interact with UAV and satellite features to establish cross-view target correspondence. The resulting tokens are finally used to regress the target location on the satellite image. 

Despite these advances, existing CVOGL approaches remain passive. The query image is fixed before localization and the model cannot decide whether or where to acquire additional observations. ActiveGeo extends this formulation to active perception, requiring a mobile agent to sequentially select new viewpoints and determine when to stop.

\section{Problem Formulation}
\label{sec:method_overview}

\textbf{From passive CVOGL to ActiveGeo.}
Cross-view object geo-localization (CVOGL) assumes a fixed query observation. In the UAV-based setting considered in this work, given a satellite reference image $I^{s}$, a query UAV image $I_{v_1}$, and a prompt $p$ specifying the target object, the model directly predicts its location $\hat{b}$ as
\begin{equation}
    \hat{b}=f(I^{s}, I_{v_1}, p),
\end{equation}
where $\hat{b}$ denotes the predicted target box. Since existing CVOGL relies entirely on the fixed query view $I_{v_1}$, it cannot acquire additional observations to resolve ambiguities and improve localization when the initial observation is unfavorable.

In contrast, ActiveGeo allows a UAV agent to sequentially acquire observations from new viewpoints. Let $\mathcal{V}=\{v_1,\ldots,v_M\}$ denote the candidate viewpoints. Each episode starts from an initial view $I_{v_1}$ and a target prompt $p$. The prompt is provided only with the initial observation, while subsequent views must be associated with the same target without repeated prompting.

At step $t$, let $\tilde{v}_t$ denote the viewpoint acquired, with $\tilde{v}_1=v_1$. The observation history $\mathcal{H}_t$ is
\begin{equation}
    \mathcal{H}_t=
    \{(I_{\tilde{v}_1},p),I_{\tilde{v}_2},\ldots,I_{\tilde{v}_t}\},
\end{equation}
from which the model predicts the current localization $\hat{b}_t$ at step $t$
\begin{equation}
    \hat{b}_t=f(I^{s},\mathcal{H}_t).
\end{equation}
The agent then decides its action $d_t $ about whether to stop or continue,
where $d_t \in \{\textsc{Stop}, \textsc{Continue}\}$. If it continues, it selects an unseen viewpoint $\tilde{v}_{t+1} \in \mathcal{V}\setminus\mathcal{V}^{\mathrm{obs}}_t,$
where $\mathcal{V}^{\mathrm{obs}}_t$ denotes the viewpoints already acquired. The UAV then moves to $\tilde{v}_{t+1}$ and obtains the corresponding observation, which is appended to $\mathcal{H}_t$. If it stops, the current prediction $\hat{b}_t$ is returned as the final localization result.

\begin{figure}[!htb]
    \centering
    \includegraphics[width=\linewidth]{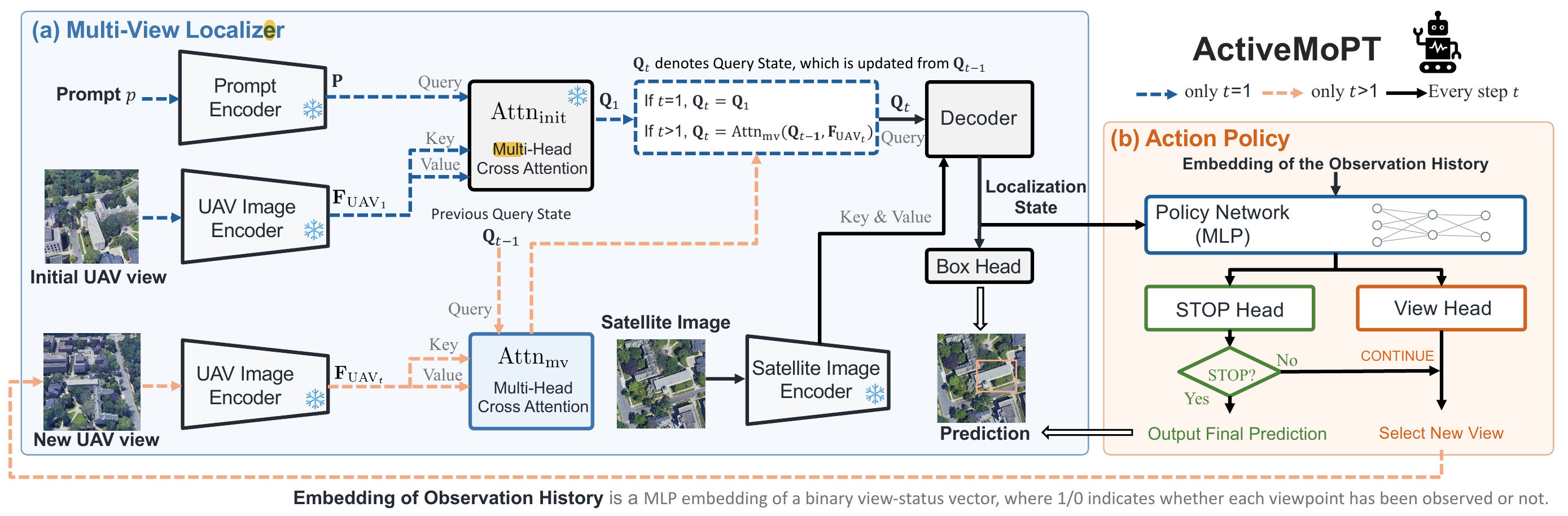}
    \vspace{-0.5cm}
    \caption{
    Overall architecture of ActiveMoPT, an agent system for ActiveGeo that performs localization, viewpoint selection, and stopping. ActiveMoPT is initialized from MoPT~\citep{mopt}. At $t=1$, before acquiring additional UAV views, its localizer is identical to MoPT.
    (a) The multi-view localizer updates the query state with newly acquired UAV views and predicts the target location on the satellite image. During Multi-View Prompt-Preserving Adaptation (Sec.~\ref{sec:method_multiview}), the encoders and $\operatorname{Attn}_{\mathrm{init}}$ are frozen.
    (b) The action policy decides whether to stop or acquire a new UAV view. Candidate UAV images are unavailable before selection, and the View Head selects among unobserved viewpoint IDs. Please refer to zoomed-in view for better visualization.
    }
    \label{fig:architecture}
    \vspace{-0.2cm}
\end{figure}

Notably, the visual content of an unseen viewpoint is unavailable before it is selected. The agent must therefore infer where to observe next from its current localization state and observation history. ActiveGeo further seeks accurate localization with as few additional observations as necessary, requiring the agent to jointly determine \emph{where to observe} and \emph{when to stop}.

\section{Methodology}
\label{sec: Methodology}

We build ActiveMoPT upon MoPT~\citep{mopt}, whose prompt tokens mediate UAV-satellite interaction and provide a natural foundation for preserving the initially specified target across subsequent unprompted UAV views.

As shown in Fig. \ref{fig:architecture}, ActiveMoPT consists of a multi-view localizer and an action policy. The localizer progressively integrates newly acquired UAV views and updates the localization of the target specified in the initial prompt, while the policy decides whether to stop and, if not, which unseen viewpoint to acquire next. ActiveMoPT is trained in three stages. (1) \emph{Multi-View Prompt-Preserving Adaptation} trains the localizer to integrate additional views without repeated prompting. (2) With the localizer frozen, \emph{Trajectory-Guided Policy Initialization} trains the policy from oracle trajectories to learn initial stopping and viewpoint selection behavior. (3) \emph{Cost-Aware Policy Refinement} further optimizes the policy through closed-loop interaction, balancing performance improvement against view acquisition cost. Together, these stages turn ActiveMoPT into an ActiveGeo agent that can continuously update its localization, actively acquire informative viewpoints, and stop once sufficient cross-view information has been collected.

\subsection{Multi-View Prompt-Preserving Adaptation}
\label{sec:method_multiview}

To build the foundation of ActiveGeo, ActiveMoPT must first be able to accumulate evidence from newly acquired views while preserving the target specified by the initial prompt. Since MoPT~\citep{mopt} only localizes the target from a single prompted UAV view, we adapt it to handle a sequence of UAV observations where only the first view is prompted. The adaptation needs to address two challenges, \emph{i.e.}, (1) maintaining the target specified by the initial prompt when incorporating new UAV views, and (2) effectively improving localization with newly acquired views while avoiding performance drops caused by uninformative observations.

\textbf{Prompt-Preserving Query Update.}
In MoPT~\citep{mopt}, the prompt tokens attend to the initial UAV view and become query tokens carrying information about the prompted target. We therefore reuse these tokens for each newly acquired view, allowing new observations to update the same tokens without requiring another prompt.

Specifically, as shown in Fig. \ref{fig:architecture}, let $\mathbf{P}$ denote the tokens encoded from the initial prompt and $\mathbf{F}_{\mathrm{UAV}_t}$ the UAV features at step $t$. Using the original Prompt-UAV cross attention $\operatorname{Attn}_{\mathrm{init}}$ in MoPT, we obtain the initial query tokens $\mathbf{Q}_{1}$
\begin{equation}
\mathbf{Q}_{1} =
\operatorname{Attn}_{\mathrm{init}}
\left(
\mathbf{P},\,
\mathbf{F}_{\mathrm{UAV}_1}
\right)
\label{eq:init_query_update}
\end{equation}
where $\mathbf{P}$ serves as queries and $\mathbf{F}_{\mathrm{UAV}_1}$ as keys and values. For each subsequent UAV view, we use a shared multi-view cross attention module $\operatorname{Attn}_{\mathrm{mv}}$ to update the query tokens $\mathbf{Q}_{t}$
\begin{equation}
\mathbf{Q}_{t} =
\operatorname{Attn}_{\mathrm{mv}}
\left(
\mathbf{Q}_{t-1},\,
\mathbf{F}_{\mathrm{UAV}_t}
\right),
\qquad t=2,\ldots,T,
\label{eq:recurrent_prompt}
\end{equation}
where $\mathbf{Q}_{t-1}$ serves as queries and $\mathbf{F}_{\mathrm{UAV}_t}$ as keys and values.

\textbf{Preserve-Improve Loss.}
Sequentially adding UAV views does not guarantee a better localization, since a newly acquired view may provide either useful or misleading information. We therefore expect each view update to improve the current prediction when useful, while avoiding causing performance degradation. 

Specifically, for sample $i$, let $u_1^i$ and $u_T^i$ denote the IoUs between predicted boxes and ground truth boxes obtained from the initial UAV view and after multi-view updating, respectively. Given a localization success threshold $\tau$, we define
\begin{equation}
\mathcal{L}_{\mathrm{PI}}^{i}
=
\begin{cases}
\max\left(0,\tau-u_T^i\right),
& u_1^i \ge \tau, \\[6pt]
\max\left(0,u_1^i-u_T^i\right)
+
\max\left(0,\tau-u_T^i\right),
& u_1^i < \tau.
\end{cases}
\label{eq:pi_loss}
\end{equation}
For an initially successful prediction ($u_1^i \ge \tau$), the first term penalizes the updated result only if it falls below the success threshold. For an initially failed prediction ($u_1^i < \tau$), the latter two terms encourage the updated IoU to exceed the initial IoU and further reach the success threshold.

Finally, the multi-view localizer is trained with
\begin{equation}
\mathcal{L}_{\mathrm{MV}}
=
\mathcal{L}_{\mathrm{Det}}^{(1)}
+
\mathcal{L}_{\mathrm{Det}}^{(T)}
+
\mathcal{L}_{\mathrm{PI}},
\label{eq:multiview_objective}
\end{equation}
where $\mathcal{L}_{\mathrm{Det}}^{(1)}$ and $\mathcal{L}_{\mathrm{Det}}^{(T)}$ are DETR-style detection losses~\citep{DETR} and supervise the initial and final localization results, respectively. The $\mathcal{L}_{\mathrm{Det}}$ optimizes localization accuracy, while $\mathcal{L}_{\mathrm{PI}}$ constrains how the prediction changes after additional views are included.

\subsection{Trajectory-Guided Policy Initialization}
\label{sec:method_sft}

\begin{figure}[!htb]
    \centering
    \includegraphics[width=\linewidth]{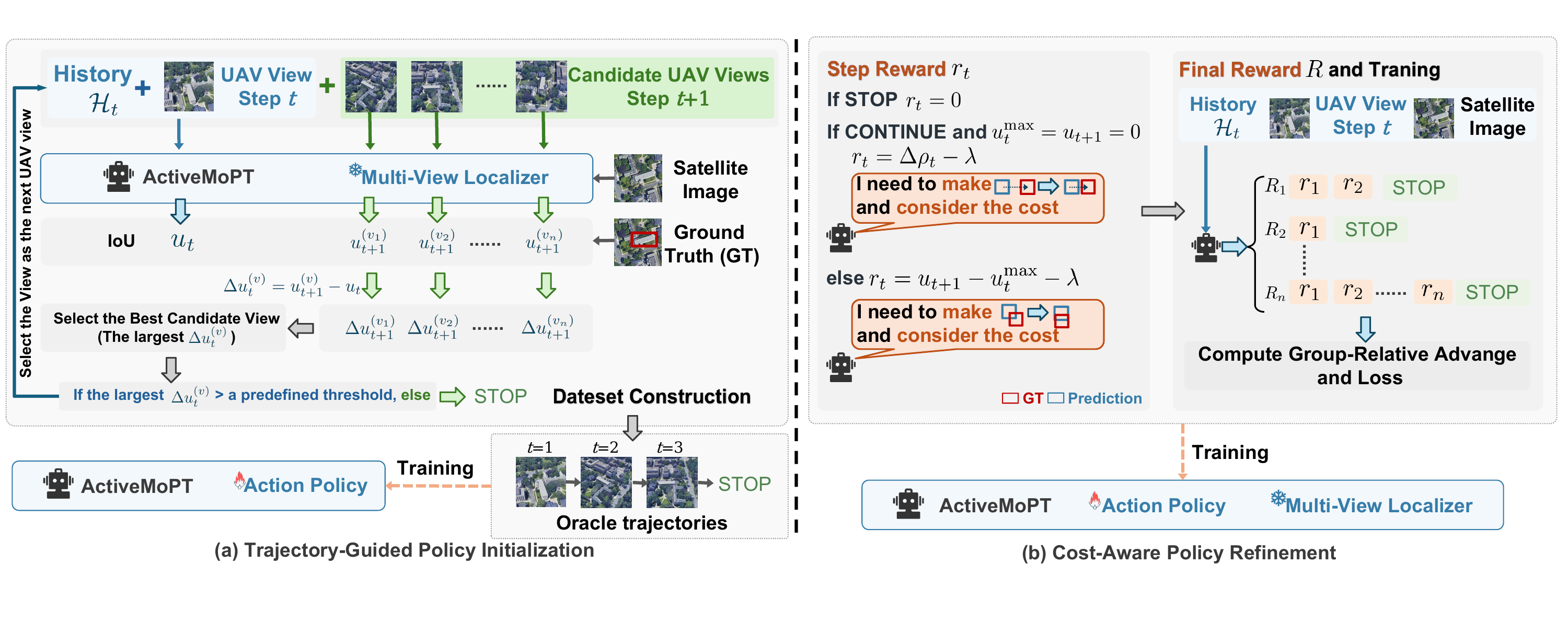}
    \vspace{-0.5cm}
    \caption{
    Training stages of the ActiveMoPT action policy.
    (a) Trajectory-Guided Policy Initialization. We construct greedy oracle trajectories to supervise the action policy during SFT.  (b) Cost-Aware Policy Refinement. Based on GRPO~\citep{GRPO}, the action policy is further optimized in closed-loop interaction. A gain-cost reward evaluates localization improvement while eliminating unnecessary observations. Please refer to zoomed-in view for better visualization.
    }
    \label{fig:sft_rl}
    \vspace{-0.2cm}
\end{figure}

With the multi-view localizer in place, ActiveMoPT still needs to learn \emph{when to stop} and \emph{which view to acquire next}. Therefore, we introduce a supervised fine-tuning (SFT) stage (Fig. \ref{fig:sft_rl}(a)) to initialize the action behavior, which provides a cold start and equips the model with a basic awareness of the action policy, ensuring the subsequent reinforcement learning (RL) explores within a meaningful policy space~\citep{AdaThinkDrive, DriveAgentR1}. 

\textbf{Construction of the oracle trajectory dataset.}
During training, all candidate UAV views and the ground-truth box are available for each scene. We use this information offline to construct supervised trajectories by evaluating candidate viewpoints and assigning an action at each step.

Given the current observation history $\mathcal{H}_t$, the frozen localizer produces a prediction $\hat{b}_t$. Its IoU $u_t$ with the ground-truth box $b^\ast$ is
\begin{equation}
u_t=\operatorname{IoU}(\hat{b}_t,b^\ast).
\end{equation}
For each unseen viewpoint $v$, we evaluate the frozen localizer using $\mathcal{H}_t$ together with $I_v$, and denote the resulting IoU by $u_{t+1}^{(v)}$. The localization gain $\Delta u_t^{(v)}$ of this candidate is
\begin{equation}
\Delta u_t^{(v)}=u_{t+1}^{(v)}-u_t.
\label{eq:oracle_gain}
\end{equation}
If the largest gain exceeds a predefined threshold, we assign \textsc{Continue} and the corresponding viewpoint as the action labels. The selected view is then added to the observation history, and the process repeats. Otherwise, we assign \textsc{Stop}. The resulting sequence of observation histories and action labels forms the oracle trajectory dataset for SFT.

\textbf{Policy learning in SFT.}
During SFT, we freeze the localizer of ActiveMoPT, while the generated trajectories provide supervision for the two policy heads. The stop head learns whether to terminate the episode, while the view head learns which unseen viewpoint to select when continuing. The training objective $\mathcal{L}_{\mathrm{SFT}}$ is
\begin{equation}
\mathcal{L}_{\mathrm{SFT}}
=
\mathcal{L}_{\mathrm{stop}}
+
\mathcal{L}_{\mathrm{view}},
\label{eq:sft_loss}
\end{equation}
where $\mathcal{L}_{\mathrm{stop}}$ is binary cross-entropy and $\mathcal{L}_{\mathrm{view}}$ is cross-entropy. At step $t$, acquired and unavailable viewpoints are masked during view selection.

\subsection{Cost-Aware Policy Refinement}
\label{sec:method_grpo}

The SFT policy learns initial viewpoint selection and stopping behavior by imitating greedy oracle trajectories. However, the oracle is constructed from localization gains and does not explicitly account for the cost of acquiring additional views. We therefore further refine the policy through closed-loop RL to learn the trade-off between localization improvement and observation cost. During this stage, the localizer remains frozen, and only the action policy is optimized using group relative policy optimization (GRPO)~\citep{GRPO}.

\textbf{Gain-cost reward.}
As shown in Fig. \ref{fig:sft_rl}(b), we reward performance improvement while charging for each additional UAV view. For each trajectory, let $u_t$ denote the localization IoU after step $t$, and let
\begin{equation}
u_t^{\max}=\max_{k\leq t}u_k
\end{equation}
denote the best IoU obtained so far at step $t$. The step reward $r_t$ is defined as
\begin{equation}
r_t=
\begin{cases}
0,
& d_t=\textsc{Stop}, \\[4pt]
\Delta \rho_t-\lambda,
& d_t=\textsc{Continue},
\quad u_t^{\max}=0,\;u_{t+1}=0, \\[4pt]
(u_{t+1}-u_t^{\max})-\lambda,
& \text{otherwise},
\end{cases}
\label{eq:gain_cost_reward}
\end{equation}
where $\lambda$ is the cost of acquiring one additional view. The IoU term rewards improvement over $u_t^{\max}$ and penalizes any drops below that level. When both $u_t^{\max}$ and $u_{t+1}$ are zero, IoU cannot indicate whether the new view improves the prediction. We therefore use the change in center distance as the reward signal in this case. Let $\rho_t$ denote the normalized center distance between the predicted and ground-truth boxes, and let $\rho_t^{\min}=\min_{k\leq t}\rho_k$ be the smallest distance obtained so far. We define $\Delta \rho_t=\rho_t^{\min}-\rho_{t+1}$, which is positive when the new prediction moves closer to the target.

To account for the rewards from subsequent decisions, the final reward $R$ is computed from each step to the end of the trajectory,
\begin{equation}
R=\sum_{k=t}^{T-1}r_k,
\label{eq:return_to_go}
\end{equation}
where $T$ is the number of observations acquired before termination.

\textbf{Policy optimization in RL.}
For each initial episode, we compare $R$ of sampled trajectories and standardize them to obtain the group-relative advantage $A$. The action policy is then optimized using the clipped GRPO objective $\mathcal{L}_{\mathrm{RL}}$~\citep{GRPO}
\begin{equation}
\mathcal{L}_{\mathrm{RL}}
=
-\mathbb{E}_{t}
\left[
\min\left(
\omega_t A,\,
\operatorname{clip}(\omega_t,1-\epsilon_c,1+\epsilon_c)A
\right)
\right]
+
\beta\,\widehat{\mathrm{KL}}
\left(\pi_\phi\,\|\,\pi_{\mathrm{SFT}}\right),
\label{eq:grpo_loss}
\end{equation}
where
$\omega_t=\pi_\phi/\pi_{\mathrm{old}}$
is the policy probability ratio, $\epsilon_c$ is the clipping radius, and $\beta$ controls the sampled KL penalty to the frozen SFT policy $\pi_{\mathrm{SFT}}$.

\section{Experiments}
\label{sec:experiments}

\subsection{Datasets and Evaluation Protocol}
\label{sec:datasets}

\textbf{MoP-UAV.}
We follow the MoP-UAV split~\citep{mopt}, which contains 598 training scenes and 100 test scenes, and evaluate all 17{,}447 target episodes in the test split. Each episode starts from one prompted UAV image. ActiveMoPT may then select an unseen UAV view or output \textsc{Stop}, while reusing the point (\prompttag{P}), box (\prompttag{B}), and/or language (\prompttag{L}) prompt provided with the initial observation.

\textbf{ActiveGeo-858.}
To evaluate zero-shot generalization beyond MoP-UAV~\citep{mopt}, we construct ActiveGeo-858 from the GeoText-1652 test split~\citep{university-1652}. It contains 858 scenes that do not overlap with the MoP-UAV~\citep{mopt}. Experts annotate two visually identifiable targets in each scene, yielding 1{,}716 evaluation episodes. Each target is annotated with a point (\prompttag{P}) and a box (\prompttag{B}) prompt. No ActiveGeo-858 image is used for training, checkpoint selection, or hyperparameter tuning.

\textbf{Evaluation protocol.}
Every episode starts from one initial UAV observation. ActiveMoPT then selects additional views and returns the prediction after selecting \textsc{Stop} or reaching a 7-view limit, whereas passive approaches~\citep{DetGeo, ReCOT, TROGeo, mopt, OCGNet} use only the initial UAV view. Ground-truth IoU is never available to the policy during evaluation.

\subsection{Evaluation Metrics}
\label{sec:metrics}

\textbf{Localization accuracy.}
We report mean intersection over union (mIoU) and Acc@$\tau$~\citep{DetGeo}, the fraction of final predictions whose IoU is at least $\tau$, for $\tau\in\{0.25,0.50\}$.

\textbf{Observation efficiency.}
We report Avg.\ Views, the average number of UAV observations used per episode, including the mandatory initial image. A lower Avg.\ Views indicates that the agent reaches its final prediction with fewer additional observations.

\subsection{Implementation Details}
\label{sec:implementation}

Please refer to the appendix.

\subsection{Comparison Results}
\label{sec:comparison}

\textbf{Quantitative results.} Table~\ref{tab:compact_main} compares ActiveMoPT with previous CVOGL approaches on both MoP-UAV~\citep{mopt} and ActiveGeo-858 dataset. 

(1) \emph{Active viewpoint selection efficiently enhances localization performance.}
Under the point-prompt (\prompttag{P}) setting, ActiveMoPT achieves the best localization results on both benchmarks. On MoP-UAV~\citep{mopt} dataset, it improves the MoPT from 0.368 to 0.384 mIoU while using only 1.291 views on average. The improvement can also be observed in Table~\ref{tab:compact_main}(b) and Table~\ref{tab:compact_main}(c). These results show the efficiency of ActiveMoPT. 

(2) \emph{ActiveMoPT generalizes well to unseen scenes.} On ActiveGeo-858, ActiveMoPT consistently improves MoPT (Table~\ref{tab:compact_main}(b)). Since no image from ActiveGeo-858 is used during checkpoint selection in training, these results show that the learned multi-view localization and action policy transfer well beyond MoP-UAV~\citep{mopt} dataset.

\textbf{Qualitative results.}
Please refer to the appendix.

\begin{table}[t]
    \centering
    \caption{Comparison on MoP-UAV~\citep{mopt} and ActiveGeo-858. (a) Comparisons of all methods under the point-prompt protocol. (b) Comparisons under \prompttag{B} and \prompttag{P+B} prompts on both benchmarks. (c) Comparisons of different prompt combinations on MoP-UAV~\citep{mopt} dataset since ActiveGeo-858 contains no language annotations.}
    \label{tab:compact_main}
    \scriptsize
    \setlength{\tabcolsep}{2.6pt}
    \renewcommand{\arraystretch}{1.04}
    \begin{tabularx}{\linewidth}{@{}Xcccccccc@{}}
        \toprule
        & \multicolumn{4}{c}{\textbf{MoP-UAV~\citep{mopt}}} & \multicolumn{4}{c}{\textbf{ActiveGeo-858}} \\
        \cmidrule(lr){2-5}\cmidrule(l){6-9}
        Method & mIoU & Acc@.25 & Acc@.50 & Avg. Views & mIoU & Acc@.25 & Acc@.50 & Avg. Views \\
        \midrule

        \multicolumn{9}{@{}l}{\textbf{(a) Point prompt (\prompttag{P})}} \\
        \midrule
        DetGeo~\citep{DetGeo} & 0.327 & 0.483 & 0.321 & 1.000 & 0.287 & 0.487 & 0.270 & 1.000 \\
        OCGNet~\citep{OCGNet} & 0.335 & 0.497 & 0.337 & 1.000 & 0.255 & 0.432 & 0.210 & 1.000 \\
        ReCOT~\citep{ReCOT} & 0.318 & 0.583 & 0.280 & 1.000 & 0.367 & 0.631 & 0.346 & 1.000 \\
        TROGeo~\citep{TROGeo} & 0.314 & 0.586 & 0.265 & 1.000 & 0.293 & 0.536 & 0.213 & 1.000 \\
        MoPT~\citep{mopt} & \second{0.368} & \second{0.598} & \second{0.405} & 1.000 & \second{0.381} & \second{0.640} & \second{0.379} & 1.000 \\
        \rowcolor{oursbg}
        ActiveMoPT (\textbf{Ours}) & \best{0.384} & \best{0.611} & \best{0.420} & 1.291 &
        \best{0.391} & \best{0.643} & \best{0.409} & 1.195 \\
        \midrule

        \multicolumn{9}{@{}l}{\textbf{(b) Comparison with spatial prompts (\prompttag{B} and \prompttag{P+B})}} \\
        \addlinespace[1pt]
        \multicolumn{9}{@{}l}{\prompttag{B}} \\
        MoPT~\citep{mopt} & \second{0.373} & \second{0.617} & \second{0.407} & 1.000 &
        \second{0.386} & \second{0.657} & \second{0.405} & 1.000 \\
        \rowcolor{oursbg}
        ActiveMoPT (\textbf{Ours}) & \best{0.389} & \best{0.630} & \best{0.432} & 1.384 &
        \best{0.414} & \best{0.689} & \best{0.451} & 1.429 \\

        \addlinespace[1pt]
        \multicolumn{9}{@{}l}{\prompttag{P+B}} \\
        MoPT~\citep{mopt} & \second{0.375} & \second{0.619} & \second{0.405} & 1.000 &
        \second{0.394} & \second{0.672} & \second{0.421} & 1.000 \\
        \rowcolor{oursbg}
        ActiveMoPT (\textbf{Ours}) & \best{0.395} & \best{0.639} & \best{0.434} & 1.403 &
        \best{0.422} & \best{0.696} & \best{0.476} & 1.371 \\
        \midrule

        \multicolumn{9}{@{}l}{\textbf{(c) comparisons of different prompt combinations on MoP-UAV~\citep{mopt}}} \\
        \addlinespace[1pt]
        & \multicolumn{4}{c}{\textbf{MoPT}~\citep{mopt}}
        & \multicolumn{4}{c}{\cellcolor{oursbg}\textbf{ActiveMoPT (Ours)}} \\
        \cmidrule(lr){2-5}\cmidrule(l){6-9}
        Prompt modality & mIoU & Acc@.25 & Acc@.50 & Avg. Views &
        mIoU & Acc@.25 & Acc@.50 & Avg. Views \\
        \midrule
        \prompttag{L}
        & \best{0.365} & \best{0.611} & \second{0.405} & 1.000
        & \cellcolor{oursbg}\best{0.365}
        & \cellcolor{oursbg}\second{0.600}
        & \cellcolor{oursbg}\best{0.407}
        & \cellcolor{oursbg}1.478 \\

        \prompttag{P+L}
        & \second{0.378} & \second{0.620} & \second{0.430} & 1.000
        & \cellcolor{oursbg}\best{0.390}
        & \cellcolor{oursbg}\best{0.626}
        & \cellcolor{oursbg}\best{0.448}
        & \cellcolor{oursbg}1.417 \\

        \prompttag{B+L}
        & \second{0.385} & \second{0.628} & \second{0.439} & 1.000
        & \cellcolor{oursbg}\best{0.407}
        & \cellcolor{oursbg}\best{0.649}
        & \cellcolor{oursbg}\best{0.472}
        & \cellcolor{oursbg}1.452 \\

        \prompttag{P+B+L}
        & \second{0.390} & \second{0.639} & \second{0.436} & 1.000
        & \cellcolor{oursbg}\best{0.414}
        & \cellcolor{oursbg}\best{0.659}
        & \cellcolor{oursbg}\best{0.478}
        & \cellcolor{oursbg}1.450 \\
        \bottomrule
    \end{tabularx}

    \parbox{\linewidth}{\raggedright\scriptsize
    ReCOT~\citep{ReCOT} uses its selected checkpoint at 6 recurrent steps. Best and second-best localization results of part (a), (b), and (c) are shown in \textcolor{bestred}{\textbf{bold red}} and \textcolor{secondblue}{blue}, respectively.}
    \vspace{-0.5cm}
    
\end{table}

\subsection{Ablation Study}
\label{sec:ablation}

\begin{table*}[t]
    \centering
    \caption{Ablations on MoP-UAV~\citep{mopt} dataset using prompt \prompttag{P+B+L}. (a) Ablation of the components in Multi-View Prompt-Preserving Adaptation stage (b) Ablation of the SFT and RL stage. ``Random" uses a sampled number of 1-4 UAV views, while ``Adaptive" lets the learned active action policy determine when to stop, up to the maximum number of UAV views.}
    \label{tab:compact_ablation}
    \scriptsize
    \setlength{\tabcolsep}{4pt}
    \renewcommand{\arraystretch}{1.12}
    \begin{tabularx}{\textwidth}{@{}Xccccc@{}}
        \toprule
        Configuration & Input UAV Views & mIoU & Acc@.25 & Acc@.50 & Avg. Views Used \\
        \midrule
        \multicolumn{6}{@{}l}{\textbf{(a) Multi-View Prompt-Preserving Adaptation}} \\
        \midrule
        \multirow{2}{*}{\shortstack[l]{with  $\operatorname{Attn}_{\mathrm{init}}$ and $\mathcal{L}_{\mathrm{PI}}$}}
        & 1 & 0.392 & 0.637 & 0.444 & 1.000 \\
        & 2 & 0.155 & 0.255 & 0.051 & 2.000 \\
        \midrule
        \multirow{5}{*}{with $\operatorname{Attn}_{\mathrm{init}}$ and $\operatorname{Attn}_{\mathrm{mv}}$}
        & 1 & 0.388 & 0.637 & 0.437 & 1.000 \\
        & 2 & 0.395 & \best{0.640} & 0.448 & 2.000 \\
        & 3 & 0.395 & 0.633 & 0.449 & 3.000 \\
        & 4 & 0.396 & 0.633 & 0.448 & 4.000 \\
        & Random & 0.393 & 0.637 & 0.446 & 2.500 \\
        \midrule
        

        \rowcolor{oursbg}
        & 1 & 0.389 & 0.632 & 0.440 & 1.000 \\
        \rowcolor{oursbg}
        & 2 & 0.398 & \best{0.640} & 0.455 & 2.000 \\
        \rowcolor{oursbg}
        & 3 & \second{0.399} & \second{0.639} & \best{0.461} & 3.000 \\
        \rowcolor{oursbg}
        & 4 & \best{0.400} & \second{0.639} & \second{0.459} & 4.000 \\
        \rowcolor{oursbg}

        \multirow{-5}{*}{with $\operatorname{Attn}_{\mathrm{init}}$, $\operatorname{Attn}_{\mathrm{mv}}$, and $\mathcal{L}_{\mathrm{PI}}$}
        & Random & 0.396 & 0.636 & 0.454 & 2.500 \\
        
        \midrule
        \multicolumn{6}{@{}l}{\textbf{(b) Acquisition policy}} \\
        \midrule
        Passive Policy (MoPT~\citep{mopt} Baseline)
        & 1 & 0.390 & 0.639 & 0.436 & 1.000 \\
        Active Policy (ActiveMoPT, SFT)
        & Adaptive & \second{0.411} & \second{0.657} & \second{0.474} & 1.886 \\
        \rowcolor{oursbg}
        Active Policy (ActiveMoPT, SFT+GRPO)
        & Adaptive & \best{0.414} & \best{0.659} & \best{0.478} & 1.450 \\
        \bottomrule
    \end{tabularx}
    
    \parbox{\linewidth}{\raggedright\scriptsize
    For the 1-4 and Random settings, the initial view is always included, and the remaining views are randomly sampled. Best and second-best localization results of parts (a) and (b) are shown in \textcolor{bestred}{\textbf{bold red}} and \textcolor{secondblue}{blue}, respectively.}
    \vspace{-0.5cm}
\end{table*}

\begin{table*}[t]
    \centering
    \caption{Ablations of key hyperparameter on MoP-UAV~\citep{mopt} dataset using prompt \prompttag{P+B+L}. (a) Ablations of the success thresholds $\tau$ in $\mathcal{L}_{\text{PI}}$. (b) Ablations of the  acquisition cost $\lambda$ in $r_{t}$. ``Random" uses a sampled number of 1-4 UAV views, while ``Adaptive" lets the learned active action policy determine when to stop, up to the maximum number of UAV views.}
    \label{tab:hyperparameter_ablation}
    \scriptsize
    \setlength{\tabcolsep}{4pt}
    \renewcommand{\arraystretch}{1.10}
    \begin{tabularx}{\textwidth}{@{}Xcccc@{}}
        \toprule
        Configuration & Input UAV Views & mIoU & Acc@.50 & Avg. Views Used \\
        \midrule
        \multicolumn{5}{@{}l}{\textbf{(a) Threshold $\tau$} in $\mathcal{L}_{\text{PI}}$} \\
        \midrule
        \multirow{5}{*}{$\tau=0.3$}
        & 1 & 0.389 & 0.439 & 1.000 \\
        & 2 & 0.393 & 0.444 & 2.000 \\
        & 3 & 0.388 & 0.438 & 3.000 \\
        & 4 & 0.386 & 0.434 & 4.000 \\
        & Random & 0.389 & 0.438 & 2.500 \\
        
        \midrule
        \rowcolor{oursbg}
        & 1 & 0.389 & 0.440 & 1.000 \\
        \rowcolor{oursbg}
        & 2 & 0.398 & 0.455 & 2.000 \\
        \rowcolor{oursbg}
        & 3 & \second{0.399} & \best{0.461} & 3.000 \\
        \rowcolor{oursbg}
        & 4 & \best{0.400} & \second{0.459} & 4.000 \\
        \rowcolor{oursbg}
        \multirow{-5}{*}{$\tau=0.5$}
        & Random & 0.396 & 0.454 & 2.500 \\
        
        \midrule
        \multirow{5}{*}{$\tau=0.7$}
        & 1 & 0.387 & 0.436 & 1.000 \\
        & 2 & 0.392 & 0.446 & 2.000 \\
        & 3 & 0.394 & 0.450 & 3.000 \\
        & 4 & 0.396 & 0.452 & 4.000 \\
        & Random & 0.391 & 0.444 & 2.500 \\
        \midrule
        \multicolumn{5}{@{}l}{\textbf{(b) Acquisition cost $\lambda$}} \\
        \midrule
        GRPO, $\lambda=0$
        & Adaptive & \best{0.417} & \best{0.479} & 1.933 \\
        \rowcolor{oursbg}
        GRPO, $\lambda=0.01$
        & Adaptive & \second{0.414} & \second{0.478} & 1.450 \\
        GRPO, $\lambda=0.1$
        & Adaptive & 0.405 & 0.463 & 1.000 \\
        \bottomrule
    \end{tabularx}
    
    \parbox{\linewidth}{\raggedright\scriptsize
     For the 1-4 and Random settings, the initial view is always included, and the remaining views are randomly sampled. Shaded rows indicate the selected hyperparameters. Best and second-best localization results of part (a) and (b) are shown in \textcolor{bestred}{\textbf{bold red}} and \textcolor{secondblue}{blue}, respectively.}
     \vspace{-0.5cm}
\end{table*}

\textbf{Component and Training Stage Ablation.} Table~\ref{tab:compact_ablation} evaluates the components of Multi-View Prompt-Preserving Adaptation and the learned action policy from SFT and RL stage.

(1) \emph{Multi-view adaptation enables consistent gains from additional observations} (Table~\ref{tab:compact_ablation}(a)).
With $\operatorname{Attn}_{\mathrm{init}}$, $\operatorname{Attn}_{\mathrm{mv}}$, and $\mathcal{L}_{\mathrm{PI}}$, mIoU improves from 0.389 with one view to 0.398, 0.399, and 0.400 with two, three, and four views, respectively. These results show that the adapted localizer consistently benefits from additional observations, supporting our central hypothesis that acquiring new viewpoints can improve localization. The performance reaches a plateau after 2-3 views on MoP-UAV~\citep{mopt}, further supporting our motivation to control acquisition cost.

(2) \emph{Both $\operatorname{Attn}_{\mathrm{mv}}$ and $\mathcal{L}_{\mathrm{PI}}$ contribute to performance improvement} (Table~\ref{tab:compact_ablation}(a)).
Introducing $\operatorname{Attn}_{\mathrm{mv}}$ avoids the performance drop caused by reusing $\operatorname{Attn}_{\mathrm{init}}$. Adding $\mathcal{L}_{\mathrm{PI}}$ further improves multi-view performance. These results show that $\operatorname{Attn}_{\mathrm{mv}}$ enables the model to incorporate new observations, while $\mathcal{L}_{\mathrm{PI}}$ helps maintain and enhance localization performance across views.

(3) \emph{SFT enables active acquisition, while GRPO further improves efficiency} (Table~\ref{tab:compact_ablation}(b)).
After SFT, ActiveMoPT gains the ability to actively acquire viewpoints and outperforms both the passive MoPT~\citep{mopt} baseline and the fixed/random acquisition results in Table~\ref{tab:compact_ablation}(a), demonstrating the effectiveness of the action policy. GRPO further improves mIoU to 0.414 and Acc@.50 to 0.478, while reducing the average number of views from 1.886 to 1.450. These results show that cost-aware refinement makes ActiveMoPT more efficient by reducing unnecessary observations without sacrificing localization performance.

\textbf{Hyperparameter Ablation.} (1) \emph{A moderate success threshold $\tau$ provides effective supervision for multi-view adaptation} (Table~\ref{tab:hyperparameter_ablation}(a)).
When $\tau$ is too small, low-quality predictions can already be regarded as successful, weakening the constraint of $\mathcal{L}_{\text{PI}}$ on subsequent multi-view updates. A larger $\tau$ imposes a stricter success criterion and places stronger constraints on difficult samples, resulting in smaller multi-view gains at $\tau=0.7$. In contrast, $\tau=0.5$ consistently improves localization as additional views are introduced and achieves the best overall multi-view performance. We therefore use $\tau=0.5$ to balance insufficient and overly strict supervision.

(2) \emph{Acquisition cost $\lambda$ affects the efficiency of the learned policy} (Table~\ref{tab:hyperparameter_ablation}(b)).
Setting $\lambda=0$ achieves the highest mIoU of 0.417 but uses 1.933 views on average, indicating a more aggressive action policy. Increasing $\lambda$ to 0.1 reduces the average number of views to 1.000, indicating that the policy becomes overly conservative and falls back to passive method. We therefore select $\lambda=0.01$, which achieves 0.414 mIoU and 0.478 Acc@.50 with only 1.450 views on average. This provides a favorable trade-off between localization performance and cost, making ActiveMoPT more efficient without sacrificing performance.

\section{Conclusion}
\label{sec: Conclusion}

We introduced ActiveGeo, extending CVOGL from a fixed-query task to an active setting where an agent can acquire additional viewpoints to enhance performance and decide when to stop. To address this task, we developed ActiveMoPT, which integrates new observations while preserving the initial prompt and employs a cost-aware action policy. Experiments demonstrate the effectiveness and efficiency of ActiveMoPT. ActiveGeo provides a promising direction for boosting CVOGL research.


\bibliography{iclr2026_conference}
\bibliographystyle{iclr2026_conference}

\newpage
\appendix
\section*{Appendix}

\begin{figure}[!htb]
  \centering
  \includegraphics[width=0.9\linewidth]{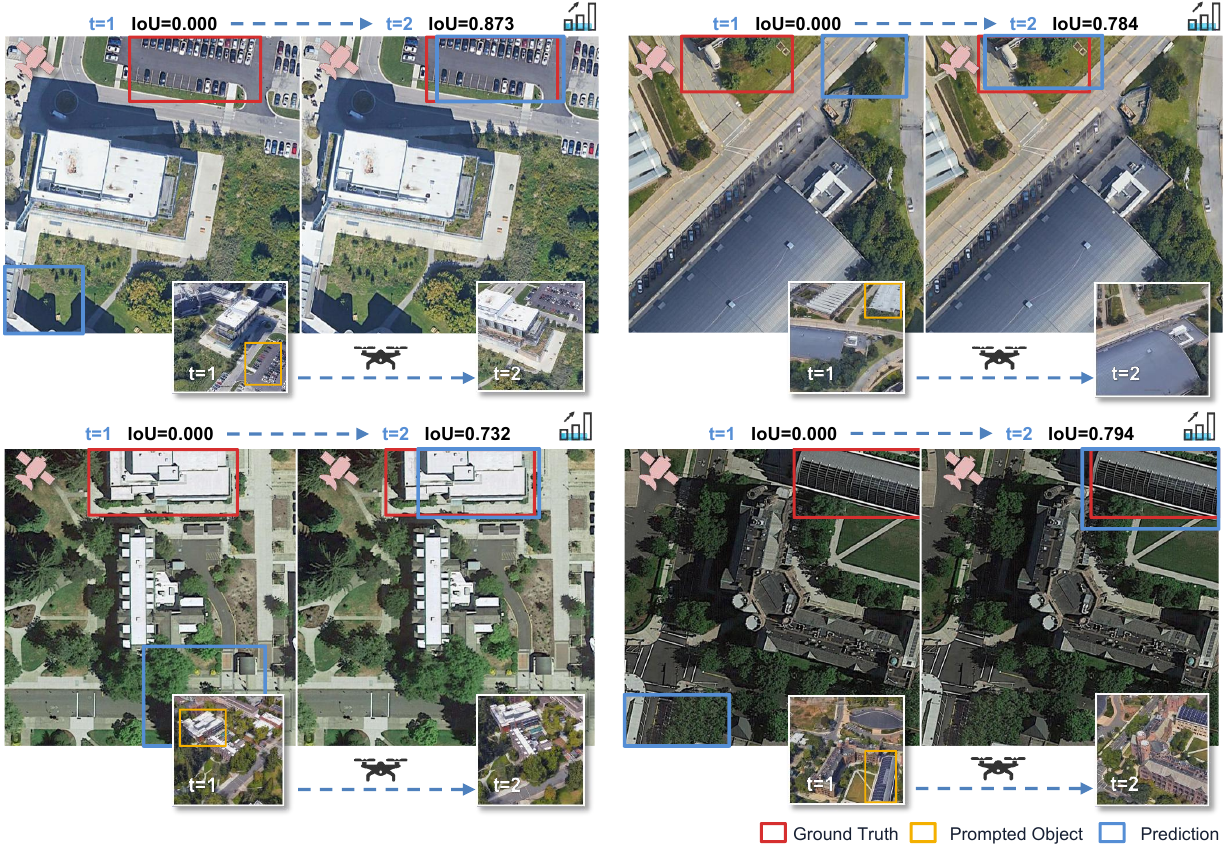}
  \vspace{-0.5cm}
   \caption{Representative results of improving first-step failed samples produced by ActiveMoPT. Notably, at $t=1$, before acquiring additional UAV views, the localizer of ActiveMoPT is identical to MoPT~\citep{mopt}. Please refer to zoomed-in view for better visualization.}
   \label{fig: vis1}
   \vspace{-0.2cm}
\end{figure}

\begin{figure}[!htb]
  \centering
  \includegraphics[width=0.7\linewidth]{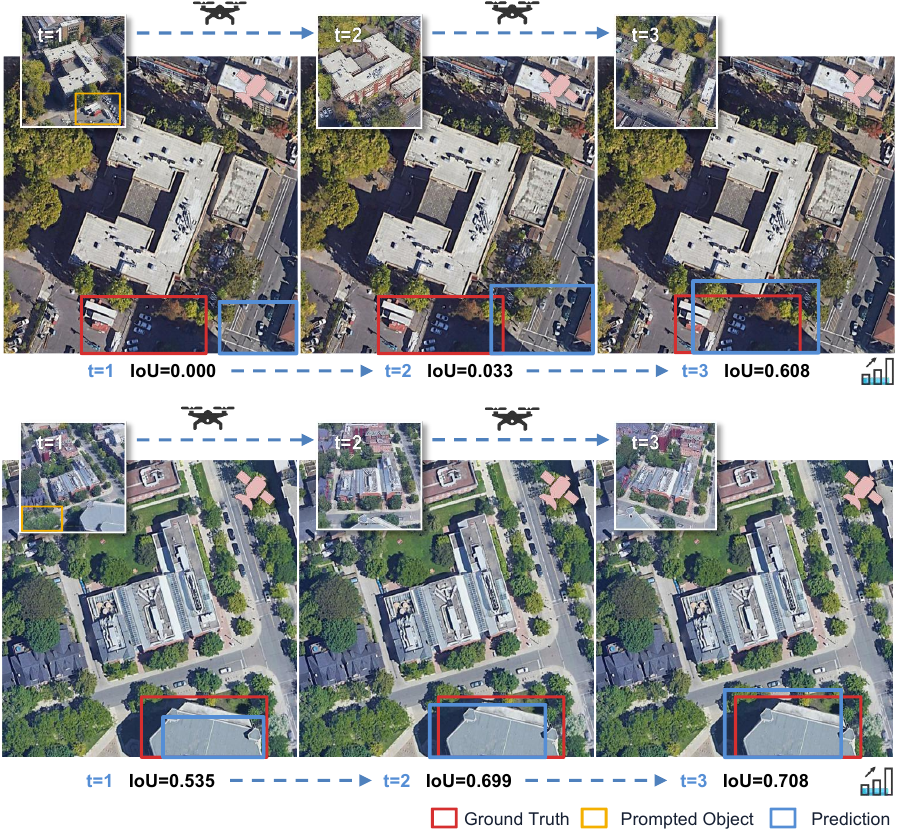}
  \vspace{-0.5cm}
   \caption{Representative results produced by ActiveMoPT. Notably, at $t=1$, before acquiring additional UAV views, the localizer of ActiveMoPT is identical to MoPT~\citep{mopt}. Please refer to zoomed-in view for better visualization.}
   \label{fig: vis2}
   \vspace{-0.2cm}
\end{figure}

\section{Implementation Details}

\textbf{Multi-view prompt-preserving adaptation.}
ActiveMoPT initializes its localizer from MoPT~\citep{mopt}, with DINOv2-Base and CLIP ViT-L/14~\citep{CLIP} as the visual and language encoders. Images are resized to $512\times512$, and the localizer uses 256-dimensional states and 100 object queries. We train this stage for 24 epochs with 2-4 UAV observations per trajectory and a learning rate of $5\times10^{-5}$. A randomly sampled non-empty subset of \prompttag{P}, \prompttag{B}, and \prompttag{L} is fixed throughout each trajectory. For $\mathcal{L}_{\mathrm{PI}}$, we set the localization success threshold to $\tau=0.5$.

\textbf{Trajectory-guided policy initialization.}
We freeze the adapted localizer and train the 770K-parameter policy for 24 SFT epochs with a learning rate of $10^{-4}$. The view head scores 54 candidate viewpoints, while the stop head predicts whether another observation is needed. During oracle trajectory generation, the oracle outputs \textsc{Stop} when the localization IoU exceeds $0.70$. Otherwise, it selects the unseen viewpoint with the largest localization gain while this gain exceeds $0.015$, and stops when no remaining viewpoint satisfies this condition. Each oracle trajectory contains at most 7 observations. The SFT checkpoint with the best closed-loop performance on MoP-UAV~\citep{mopt} is used to initialize the following GRPO stage.

\textbf{Cost-aware policy refinement.}
Starting from the selected SFT policy, we keep the localizer frozen and optimize only the policy for 2 GRPO epochs. Each episode uses groups of 32 sampled trajectories. The policy learning rate is $10^{-5}$, the clipping radius is $\epsilon_{\mathrm{c}}=0.2$, and the KL coefficient is $\beta=0.02$. We set $\lambda=0.01$. 

\textbf{Viewpoint representation.}
Following the viewpoint setting of University-1652~\citep{university-1652}, each satellite image is associated with 54 predefined UAV viewpoints. These viewpoints are sampled along three circular trajectories around the target from higher to lower elevations, resulting in 54 discrete viewpoint IDs. We therefore formulate viewpoint selection as a 54-way classification problem. Each ID corresponds to one fixed UAV viewpoint, and the view head predicts a score over the 54 IDs. At each step, viewpoints that have already been observed are masked out, and the highest-scoring remaining ID is selected when the policy decides to continue.

\section{More Experimental Results}
\subsection{Qualitative results}

(1) \emph{ActiveMoPT learns to select UAV views that are easier to match with the satellite image}(Fig.~\ref{fig: vis1}).
In these representative cases, the second UAV view selected by ActiveMoPT is often more consistent with the satellite image in object orientation and surrounding spatial layout than the initial view. This reduces the cross-view discrepancy and makes the target easier to match, allowing several first-step failures with zero IoU to be corrected after acquiring the new view. Although no explicit alignment objective is introduced, the learned policy shows an implicit tendency to select viewpoints that are easier to register with the satellite image. This indicates that ActiveMoPT does not simply acquire more observations, but actively searches for viewpoints that are more favorable for cross-view localization.

(2) \emph{ActiveMoPT progressively improves localization} (Fig.~\ref{fig: vis2}).
The examples show that newly selected views can continuously refine the localization. In the first case, the prediction remains inaccurate after the second view but is corrected after the third view, with IoU increasing from 0.000 to 0.033 and then to 0.608. In the second case, the first-step prediction is already successful, yet additional views further improve the IoU from 0.535 to 0.699 and 0.708. These examples show that different samples require different amounts of visual evidence, and that ActiveMoPT can progressively accumulate information from multiple observations to refine the target localization.

\subsection{More Ablation Results}

\begin{table}[!htb]
    \centering
    \caption{
    Ablation of the input of action policy on MoP-UAV~\citep{mopt} dataset using prompt \prompttag{P+B+L}. Best results are shown in \textcolor{bestred}{\textbf{bold red}}
    }
    \label{tab:strict_history_ablation}
    \scriptsize
    \begin{tabular}{lcccc}
    \toprule
    Input of the Action Policy
    & mIoU $\uparrow$
    & Acc@0.25 $\uparrow$
    & Acc@0.50 $\uparrow$
    & Avg. views $\downarrow$ \\
    \midrule
    w/ Embedding of the Observation History
    & \best{0.414}
    & \best{0.659}
    & \best{0.478}
    & 1.450 \\

    w/o Embedding of the Observation History
    & 0.411
    & 0.657
    & 0.470
    & \best{1.304} \\
    \bottomrule
    \end{tabular}
  \end{table}

\textbf{Effect of the observation-history embedding.}
Table~\ref{tab:strict_history_ablation} evaluates whether the action policy needs explicit information about previously observed viewpoints. The observation-history embedding (Fig.~\ref{fig:architecture}(b)) provides a binary view-status representation, indicating which candidate viewpoints have already been observed and which remain unseen. Removing this input reduces localization performance while also decreasing the average number of acquired views. This suggests that, without explicit observation history, the policy tends to stop earlier because it cannot fully distinguish how much viewpoint coverage has already been obtained from the current state alone. In contrast, providing the history embedding allows the policy to condition its stopping and viewpoint-selection decisions on the accumulated observation process, leading to more informative additional acquisitions and better localization. These results show that the observation history is not merely auxiliary state information, but an important cue for active decision making.

\begin{table}[!htb]
  \centering
  \caption{
  Ablation of $\Delta \rho_t$ in the step reward $r_t$ on MoP-UAV~\citep{mopt} dataset using \prompttag{P+B+L}. Best results are shown in \textcolor{bestred}{\textbf{bold red}}
  }
  \label{tab:zero_iou_center_reward}
  \scriptsize
  \begin{tabular}{lcccc}
  \toprule
  Reward Setting
  & mIoU $\uparrow$
  & Acc@0.25 $\uparrow$
  & Acc@0.50 $\uparrow$
  & Avg. views $\downarrow$ \\
  \midrule
  w/ $\Delta \rho_t$
  & \best{0.414}
  & \best{0.659}
  & \best{0.478}
  & 1.450 \\

  w/o $\Delta \rho_t$
  & 0.411
  & 0.654
  & 0.473
  & \best{1.235} \\
  \bottomrule
  \end{tabular}
  \end{table}

\textbf{Effect of the $\Delta \rho_t$ in $r_t$.}
Table~\ref{tab:zero_iou_center_reward} evaluates the effect of $\Delta \rho_t$ when both the current best IoU and the new IoU are zero. In this case, IoU provides no information about whether the new UAV view improves the prediction. Without the $\Delta \rho_t$ term, acquiring an additional view receives only the acquisition cost, which encourages the policy to stop earlier even when the prediction is moving toward the correct direction. This is reflected by the lower average number of views but also worse localization performance. By introducing $\Delta \rho_t$, the policy can distinguish useful progress before box overlap is achieved and continue acquiring views when they move the prediction closer to the target. The result shows that $\Delta \rho_t$ provides an important learning signal for difficult cases where IoU alone is too sparse to guide the policy.

\section{Discussion}
\label{Discussion}

\subsection{Discussion of $\mathcal{L}_{\mathrm{PI}}$}
\label{dis:pi_discussion}

We further discuss the design of the Preserve-Improve (PI) loss $\mathcal{L}_{\mathrm{PI}}$ and the role of each term in the $\mathcal{L}_{\mathrm{PI}}$. In Sec.~\ref{sec:method_multiview} the $\mathcal{L}_{\mathrm{PI}}$ of sample $i$ is defined as
\begin{equation}
\mathcal{L}_{\mathrm{PI}}^{i}
=
\begin{cases}
\max\left(0,\tau-u_T^i\right),
& u_1^i \ge \tau, \\[5pt]
\max\left(0,u_1^i-u_T^i\right)
+
\max\left(0,\tau-u_T^i\right),
& u_1^i < \tau,
\end{cases}
\label{eq:app_pi_loss}
\end{equation}
where $u_1^i$ denotes the localization IoU between predicted boxes and ground truth boxes at the first step using only the initial UAV view, and $u_T^i$ denotes the final IoU after incorporating additional views. We regard the first-step localization as successful if $u_1^i \ge \tau$, and as failed otherwise. 

The $\mathcal{L}_{\mathrm{PI}}$ treats these two cases differently. For an initially successful sample, we require the final prediction to remain above the success threshold:
\begin{equation}
\max\left(0,\tau-u_T^i\right).
\end{equation}
This prevents additional UAV views from turning an already successful localization into a failed one. If the first-step prediction fails ($u_1^i<\tau$), two terms are used:
\begin{equation}
\max\left(0,u_1^i-u_T^i\right)
+
\max\left(0,\tau-u_T^i\right).
\end{equation}
The first term $\max\left(0,u_1^i-u_T^i\right)$ becomes non-zero only when the final IoU is lower than the first-step IoU, and therefore discourages new views from making the localization worse. The second term $\max\left(0,\tau-u_T^i\right)$ remains non-zero as long as the final IoU is below $\tau$, encouraging the model to continue improving the prediction until it reaches the success threshold. For example, if a sample improves from $u_1^i=0.3$ to $u_T^i=0.4$ with $\tau=0.5$, the first term becomes zero because the localization has improved, while the second term remains active because the prediction is still unsuccessful. If the final IoU reaches $0.5$ or higher, both terms become zero.

\textbf{Weighted $\mathcal{L}_{\mathrm{PI}}$.}
To validate the contribution of these three terms, we assign a separate weight to each of them and define a weighted $\mathcal{L}_{\mathrm{PI}}$ as 
\begin{equation}
\mathcal{L}_{\mathrm{PI}}^{i}
=
\begin{cases}
w_{\mathrm{s}}\max\left(0,\tau-u_T^i\right),
& u_1^i \ge \tau, \\[5pt]
w_{\mathrm{m}}\max\left(0,u_1^i-u_T^i\right)
+
w_{\mathrm{r}}\max\left(0,\tau-u_T^i\right),
& u_1^i < \tau.
\end{cases}
\label{eq:app_pi_weighted}
\end{equation}
Here, $w_{\mathrm{s}}$ controls how strongly we keep first-step successful samples above the success threshold, $w_{\mathrm{m}}$ controls how strongly we avoid making first-step failed samples worse, and $w_{\mathrm{r}}$ controls how strongly we encourage failed samples to reach the success threshold.

Table~\ref{tab:app_pi_components} verifies these roles by changing one weight at a time from the default setting $(w_{\mathrm{s}},w_{\mathrm{m}},w_{\mathrm{r}})=(3,1,0.5)$. Removing $w_{\mathrm{m}}$ causes the largest performance drop, from 0.398 to 0.382 mIoU and from 0.457 to 0.435 Acc@.50, showing that preventing harmful multi-view updates is particularly important. Setting $w_{\mathrm{r}}=0$ also decreases performance to 0.389 mIoU and 0.445 Acc@.50, showing that merely avoiding worse predictions is insufficient. Failed samples should also be encouraged to reach the success threshold. The $w_{\mathrm{s}}$ ablation shows that protecting first-step successful predictions is also useful, while an excessively large weight can restrict the benefit of later views. Overall, the default weights provide the best balance among these three items.

\begin{table}[!htb]
    \centering
    \caption{Ablation of $(w_{\mathrm{s}},w_{\mathrm{m}},w_{\mathrm{r}})$ at $\tau=0.5$ in $\mathcal{L}_{\mathrm{PI}}$ on MoP-UAV~\citep{mopt} dataset using prompt \prompttag{P+B+L}.
    ``Mean mIoU" and ``Mean Acc@.50" and   the average of mIoU and Acc@.50 among fixed 2/3/4/Random number of UAV views. ``Random" uses a sampled number of 1-4 UAV views.
    }
    \label{tab:app_pi_components}
    \scriptsize
    \setlength{\tabcolsep}{4pt}
    \renewcommand{\arraystretch}{1.12}
    \begin{tabularx}{\linewidth}{@{}Xccccc@{}}
        \toprule
        Setting
        & $w_{\mathrm{s}}$
        & $w_{\mathrm{m}}$
        & $w_{\mathrm{r}}$
        & Mean mIoU
        & Mean Acc@.50 \\
        \midrule
        \rowcolor{oursbg}
        Default & 3 & 1 & 0.5 & \best{0.398} & \best{0.457} \\
        \midrule
        $w_{\mathrm{s}}=0$ & 0 & 1 & 0.5 & 0.396 & 0.454 \\
        $w_{\mathrm{s}}=1$ & 1 & 1 & 0.5 & 0.390 & 0.444 \\
        $w_{\mathrm{s}}=5$ & 5 & 1 & 0.5 & 0.387 & 0.439 \\
        \midrule
        $w_{\mathrm{m}}=0$   & 3 & 0   & 0.5 & 0.382 & 0.435 \\
        $w_{\mathrm{m}}=0.5$ & 3 & 0.5 & 0.5 & 0.387 & 0.442 \\
        $w_{\mathrm{m}}=2$   & 3 & 2   & 0.5 & 0.396 & 0.453 \\
        \midrule
        $w_{\mathrm{r}}=0$    & 3 & 1 & 0    & 0.389 & 0.445 \\
        $w_{\mathrm{r}}=0.25$ & 3 & 1 & 0.25 & 0.389 & 0.441 \\
        $w_{\mathrm{r}}=1$    & 3 & 1 & 1    & 0.394 & 0.443 \\
        \bottomrule
    \end{tabularx}
    \parbox{\linewidth}
    {\raggedright\scriptsize
    Shaded rows indicate the selected hyperparameters. Best results are shown in \textcolor{bestred}{\textbf{bold red}}}.
\end{table}

\subsection{Discussion of $\mathcal{L}_{\mathrm{MV}}$}

In Multi-View Prompt-Preserving Adaptation (Sec.~\ref{sec:method_multiview}), we define $\mathcal{L}_{\mathrm{MV}}$ as
\begin{equation}
\mathcal{L}_{\mathrm{MV}}
=
\mathcal{L}_{\mathrm{Det}}^{(1)}
+
\mathcal{L}_{\mathrm{Det}}^{(T)}
+
\mathcal{L}_{\mathrm{PI}},
\label{eq:app_mv_objective}
\end{equation}
where $\mathcal{L}_{\mathrm{Det}}^{(1)}$ and
$\mathcal{L}_{\mathrm{Det}}^{(T)}$ are detection losses~\citep{DETR} and supervise the localization results at the first
and final steps, respectively, and $\mathcal{L}_{\mathrm{PI}}$ constrains
how localization changes after additional UAV views are incorporated.
We further discuss two design choices in $\mathcal{L}_{\mathrm{MV}}$, \emph{i.e.},
(1) why detection supervision ($\mathcal{L}_{\mathrm{Det}}^{(1)}$ and
$\mathcal{L}_{\mathrm{Det}}^{(T)}$) and $\mathcal{L}_{\mathrm{MV}}$ are both needed, and
(2) why detection supervision ($\mathcal{L}_{\mathrm{Det}}^{(1)}$ and
$\mathcal{L}_{\mathrm{Det}}^{(T)}$) is applied only at the first ($\mathcal{L}_{\mathrm{Det}}^{(1)}$) and final steps ($\mathcal{L}_{\mathrm{Det}}^{(T)}$) rather than at every step.

\subsubsection{Why Are Detection Supervision and PI Both Needed?}
\label{app:pi_rationale}

The two losses serve different purposes.
Detection supervision ($\mathcal{L}_{\mathrm{Det}}^{(1)}$ and
$\mathcal{L}_{\mathrm{Det}}^{(T)}$) directly optimizes the predicted target box,
whereas $\mathcal{L}_{\mathrm{PI}}$ focuses on whether incorporating additional views leads to a desirable performance improvement in localization results.
For example, once the final IoU reaches the success threshold $\tau$,
the $\mathcal{L}_{\mathrm{PI}}$ becomes zero, while the detection loss can still improve
the localization performance beyond this threshold.
Conversely, $\mathcal{L}_{\mathrm{PI}}$ explicitly discourages incorporating additional UAV views from turning
a good localization into a failed one or making an initially failed
localization even worse.

The ablation in Table~\ref{tab:compact_ablation}(a) supports this
complementary design.
With both $\operatorname{Attn}_{\mathrm{init}}$ and
$\operatorname{Attn}_{\mathrm{mv}}$, adding $\mathcal{L}_{\mathrm{PI}}$
improves mIoU.
These results show that detection supervision alone can train the
localizer, while $\mathcal{L}_{\mathrm{PI}}$ further helps the model make better use of
additional views.

\subsubsection{Why Supervise Only the First and Final Steps?}
\label{app:mv_supervision}

A multi-view sequence contains several localization steps.
One straightforward choice is to apply a detection loss after every
new UAV view is incorporated.
However, we supervise only two predictions, \emph{i.e.}, the first-step prediction obtained from the initial UAV view, and the final prediction after all sampled UAV views have been incorporated. This design follows the role of the two endpoints in our setting.
The first-step loss maintains the original single-view localization
ability of MoPT~\citep{mopt}, while the final-step loss directly optimizes the result after
multi-view integration.

The intermediate states serve two roles, \emph{i.e.}, producing the current localization and carrying information for subsequent view updates. Applying a detection loss at every step directly optimizes each intermediate prediction for its current localization accuracy. However, a state that is optimal for the current step is not necessarily the most suitable state for incorporating later observations. Since the recurrent updates share parameters across steps, repeatedly optimizing all intermediate predictions may therefore limit the model's ability to organize the query state for better final multi-view integration~\citep{ReCOT}. Notably, the intermediate updates are still trained because the final loss is backpropagated through the entire recurrent update process. In addition, the number of UAV views varies across training samples, so predictions under different numbers of observations can also become final predictions and receive direct supervision.

To further validate our design, we also quantitatively compare it with supervision at every step. As shown in Table~\ref{tab:app_supervision}, supervising only the first
and final steps outperforms. This shows that directly focusing supervision on the initial and final predictions better helps the model translate additional observations into localization gains.

\begin{table}[!htb]
    \centering
    \caption{Ablation of supervision steps in $\mathcal{L}_{\text{MV}}$ of Sec.~\ref{sec:method_multiview} on MoP-UAV~\citep{mopt} dataset using prompt \prompttag{P+B+L}. We report mIoU at different fixed number of input UAV views. ``Random" uses a sampled number of 1-4 UAV views}
    \label{tab:app_supervision}
    \scriptsize
    \setlength{\tabcolsep}{4pt}
    \renewcommand{\arraystretch}{1.12}

    \begin{tabularx}{\linewidth}{@{}Xccccc@{}}
        \toprule
        Supervision steps of $\mathcal{L}_{\text{MV}}$
        & 1 view & 2 views & 3 views & 4 views & Random \\
        \midrule
        \rowcolor{oursbg}
        First and final steps
        & \best{0.389} & \best{0.398} & \best{0.399} & \best{0.400} & \best{0.396}\\
        Every step
        & 0.388 & 0.390 & 0.390 & 0.388 & 0.389 \\
        \bottomrule
    \end{tabularx}

    \vspace{2pt}
    \parbox{\linewidth}{\raggedright\scriptsize
    Shading indicates the selected supervision scheme of $\mathcal{L}_{\text{MV}}$.
    The better result is shown in
    \textcolor{bestred}{\textbf{bold red}}.}
\end{table}

\subsection{Discussion of the Predefined Threshold in SFT Dataset Construction}

During oracle trajectory construction in Sec.~\ref{sec:method_sft}, the predefined threshold determines whether an additional UAV view provides sufficient localization improvement to justify continuing the trajectory. Specifically, if the largest IoU gain among the unseen viewpoints exceeds this threshold, the corresponding view is selected. Otherwise, the trajectory stops. The threshold therefore controls how aggressively the oracle acquires additional observations.

We set this threshold to $0.015$ based on statistics of the training data. Setting it to $0$ makes the oracle overly aggressive, since even a very small positive gain leads to another view acquisition. In contrast, a larger threshold can stop the trajectory too early and discard useful additional observations. With a threshold of $0.015$, 49.32\% of the constructed trajectories obtain localization improvement, whereas increasing the threshold to $0.03$ reduces this proportion to 37.01\%. We therefore choose $0.015$ as a moderate value that retains useful multi-view trajectories without encouraging unnecessary acquisitions.

We note that a fixed threshold cannot perfectly determine the best action for every sample. This is also why SFT is used only to initialize the action policy. In the subsequent RL stage, the policy no longer follows this predefined threshold. Instead, it interacts with the frozen localizer and learns from the gain-cost reward to determine whether the expected localization improvement is worth the cost of acquiring another view. Therefore, the final policy can further adjust the trajectories learned from SFT and reduce its dependence on the manually chosen threshold.

\subsection{Discussion of the Distribution in UAV View Selection}

\begin{table*}[!htb]
    \centering
    \caption{
    Distribution of the most frequently selected UAV viewpoints at the second and third steps.
    (a) Distribution of the UAV viewpoints selected as the second observation.
    (b) Distribution of the UAV viewpoints selected as the third observation.
    }
    \label{tab:view_selection_distribution}
    \scriptsize
    \setlength{\tabcolsep}{3.2pt}
    \renewcommand{\arraystretch}{1.18}

    \textbf{(a) Distribution of the Second-View Selection}\\

    \begin{tabular*}{\textwidth}{@{\extracolsep{\fill}}lccccccccc@{}}
        \toprule
        UAV Viewpoint ID
        & 54 & 45 & 46 & 50 & 53 & 44 & 9 & 14 & 47 \\
        \midrule
        Ratio
        & 30.44\% & 9.67\% & 7.35\% & 5.67\% & 4.95\%
        & 3.70\% & 3.55\% & 3.14\% & 2.48\% \\
        \bottomrule
    \end{tabular*}
    \vspace{0.2cm}

    \textbf{(b) Distribution of the Third-View Selection}\\

    \begin{tabular*}{\textwidth}{@{\extracolsep{\fill}}lccccccccc@{}}
        \toprule
        UAV Viewpoint ID
        & 45 & 37 & 53 & 54 & 44 & 50 & 41 & 5 & 38 \\
        \midrule
        Ratio
        & 22.03\% & 15.25\% & 15.25\% & 6.78\% & 6.78\%
        & 5.08\% & 5.08\% & 5.08\% & 5.08\% \\
        \bottomrule
    \end{tabular*}
\end{table*}

Table~\ref{tab:view_selection_distribution} shows that ActiveMoPT exhibits clear but non-degenerate preferences when selecting additional UAV views. For the second observation, View 54 is selected most frequently. This is consistent with the qualitative results in Fig.~\ref{fig: vis1}, where the policy tends to favor UAV views whose orientation and spatial layout are more similar to the satellite image, making cross-view matching easier. In MoP-UAV~\citep{mopt} dataset, View 54 often provides such a favorable viewpoint, so its high selection frequency reflects a meaningful preference.

At the same time, the policy does not collapse to View 54. The remaining selections are distributed across multiple viewpoints, indicating that the preferred view depends on the current observation and localization state. A similar pattern appears at the third observation. The dominant choices shift to Views 45, 37, and 53, showing that the policy adapts its next-view selection after incorporating previous observations. 

\subsection{Limitations and Future Work}

Our current formulation considers a discrete viewpoint space, where the view head selects among predefined viewpoint IDs through classification. This setting allows us to focus on the key challenge of introducing ActiveGeo, but it does not model continuous UAV motion between viewpoints. As an initial step toward agentic CVOGL, ActiveMoPT establishes the basic interaction loop of localization, viewpoint selection, and stopping. The next step is to extend this framework to continuous trajectory control, where the UAV jointly plans its motion and acquires informative observations during navigation.

\end{document}